\documentclass[11pt, a4paper, onecolumn, copyright, goog]{google}

\usepackage{multirow}
\usepackage[authoryear, sort&compress, round]{natbib}
\uselogo{}

\title{EmbeddingGemma: Powerful and Lightweight Text Representations}

\renewcommand{\today}{}

\author[*]{Henrique Schechter Vera}
\author[*]{Sahil Dua}
\author[ \hspace{-0.3em}]{Biao Zhang}
\author[ \hspace{-0.3em}]{Daniel Salz}
\author[ \hspace{-0.3em}]{Ryan Mullins}
\author[ \hspace{-0.3em}]{Sindhu Raghuram Panyam}
\author[ \hspace{-0.3em}]{Sara Smoot}
\author[ \hspace{-0.3em}]{Iftekhar Naim}
\author[ \hspace{-0.3em}]{Joe Zou}
\author[ \hspace{-0.3em}]{Feiyang Chen}
\author[ \hspace{-0.3em}]{Daniel Cer}
\author[ \hspace{-0.3em}]{Alice Lisak}
\author[ \hspace{-0.3em}]{Min Choi}
\author[ \hspace{-0.3em}]{Lucas Gonzalez}
\author[ \hspace{-0.3em}]{Omar Sanseviero}
\author[ \hspace{-0.3em}]{Glenn Cameron}
\author[ \hspace{-0.3em}]{Ian Ballantyne}
\author[ \hspace{-0.3em}]{Kat Black}
\author[ \hspace{-0.3em}]{Kaifeng Chen}
\author[ \hspace{-0.3em}]{Weiyi Wang}
\author[ \hspace{-0.3em}]{Zhe Li}
\author[ \hspace{-0.3em}]{Gus Martins}
\author[ \hspace{-0.3em}]{Jinhyuk Lee}
\author[ \hspace{-0.3em}]{Mark Sherwood}
\author[ \hspace{-0.3em}]{Juyeong Ji}
\author[ \hspace{-0.3em}]{Renjie Wu}
\author[ \hspace{-0.3em}]{Jingxiao Zheng}
\author[ \hspace{-0.3em}]{Jyotinder Singh}
\author[ \hspace{-0.3em}]{Abheesht Sharma}
\author[ \hspace{-0.3em}]{Divyashree Sreepathihalli}
\author[ \hspace{-0.3em}]{Aashi Jain}
\author[ \hspace{-0.3em}]{Adham Elarabawy}
\author[ \hspace{-0.3em}]{AJ Co}
\author[ \hspace{-0.3em}]{Andreas Doumanoglou}
\author[ \hspace{-0.3em}]{Babak Samari}
\author[ \hspace{-0.3em}]{Ben Hora}
\author[ \hspace{-0.3em}]{Brian Potetz}
\author[ \hspace{-0.3em}]{Dahun Kim}
\author[ \hspace{-0.3em}]{Enrique Alfonseca}
\author[ \hspace{-0.3em}]{Fedor Moiseev}
\author[ \hspace{-0.3em}]{Feng Han}
\author[ \hspace{-0.3em}]{Frank Palma Gomez}
\author[ \hspace{-0.3em}]{Gustavo Hern{\'{a}}ndez {\'{A}}brego}
\author[ \hspace{-0.3em}]{Hesen Zhang}
\author[ \hspace{-0.3em}]{Hui Hui}
\author[ \hspace{-0.3em}]{Jay Han}
\author[ \hspace{-0.3em}]{Karan Gill}
\author[ \hspace{-0.3em}]{Ke Chen}
\author[ \hspace{-0.3em}]{Koert Chen}
\author[ \hspace{-0.3em}]{Madhuri Shanbhogue}
\author[ \hspace{-0.3em}]{Michael Boratko}
\author[ \hspace{-0.3em}]{Paul Suganthan}
\author[ \hspace{-0.3em}]{Sai Meher Karthik Duddu}
\author[ \hspace{-0.3em}]{Sandeep Mariserla}
\author[ \hspace{-0.3em}]{Setareh Ariafar}
\author[ \hspace{-0.3em}]{Shanfeng Zhang}
\author[ \hspace{-0.3em}]{Shijie Zhang}
\author[ \hspace{-0.3em}]{Simon Baumgartner}
\author[ \hspace{-0.3em}]{Sonam Goenka}
\author[ \hspace{-0.3em}]{Steve Qiu}
\author[ \hspace{-0.3em}]{Tanmaya Dabral}
\author[ \hspace{-0.3em}]{Trevor Walker}
\author[ \hspace{-0.3em}]{Vikram Rao}
\author[ \hspace{-0.3em}]{Waleed Khawaja}
\author[ \hspace{-0.3em}]{Wenlei Zhou}
\author[ \hspace{-0.3em}]{Xiaoqi Ren}
\author[ \hspace{-0.3em}]{Ye Xia}
\author[ \hspace{-0.3em}]{Yichang Chen}
\author[ \hspace{-0.3em}]{Yi-Ting Chen}
\author[ \hspace{-0.3em}]{Zhe Dong}
\author[ \hspace{-0.3em}]{Zhongli Ding}
\author[ \hspace{-0.3em}]{Francesco Visin}
\author[ \hspace{-0.3em}]{Ga\"{e}l Liu}
\author[ \hspace{-0.3em}]{Jiageng Zhang}
\author[ \hspace{-0.3em}]{Kathleen Kenealy}
\author[ \hspace{-0.3em}]{Michelle Casbon}
\author[ \hspace{-0.3em}]{Ravin Kumar}
\author[ \hspace{-0.3em}]{Thomas Mesnard}
\author[ \hspace{-0.3em}]{Zach Gleicher}
\author[ \hspace{-0.3em}]{Cormac Brick}
\author[ \hspace{-0.3em}]{Olivier Lacombe}
\author[ \hspace{-0.3em}]{Adam Roberts}
\author[ \hspace{-0.3em}]{Qin Yin}
\author[ \hspace{-0.3em}]{Yunhsuan Sung}
\author[ \hspace{-0.3em}]{Raphael Hoffmann}
\author[ \hspace{-0.3em}]{Tris Warkentin}
\author[ \hspace{-0.3em}]{Armand Joulin}
\author[ \hspace{-0.3em}]{Tom Duerig,}
\author[ \hspace{-0.3em}]{Mojtaba Seyedhosseini}

\correspondingauthor{hschechter@google.com}

\affil[ \hspace{-0.3em}]{EmbeddingGemma Team, Google\footnote{See Contributions and Acknowledgments section. $^*$Co-first authors; equal contributions.}}

\begin{abstract}
We introduce EmbeddingGemma, a new lightweight, open text embedding model based on the Gemma~3 language model family. Our innovative training recipe strategically captures knowledge from larger models via encoder-decoder initialization and geometric embedding distillation. We improve model robustness and expressiveness with a spread-out regularizer, and ensure generalizability by merging checkpoints from varied, optimized mixtures. Evaluated on the Massive Text Embedding Benchmark (MTEB) across multilingual, English, and code domains, EmbeddingGemma (300M) achieves state-of-the-art results. Notably, it outperforms prior top models, both proprietary and open, with fewer than 500M parameters, and provides performance comparable to models double its size, offering an exceptional performance-to-cost ratio. Remarkably, this lead persists when quantizing model weights or truncating embedding outputs. This makes EmbeddingGemma particularly well-suited for low-latency and high-throughput use cases such as on-device applications. We provide ablation studies exploring our key design choices. We release EmbeddingGemma to the community to promote further research.
\end{abstract}

\begin{document}

\maketitle

\section{Introduction}
Text embedding models map natural language to fixed-length vector representations, placing semantically similar text near each other in the embedding space. This has made them a popular choice for tasks like semantic similarity, information retrieval, and clustering. Recent research has built towards the goal of creating general-purpose models that are able to excel at many of these tasks simultaneously, attempting to develop the models themselves \citep{su2023embeddertaskinstructionfinetunedtext} and crafting benchmarks focused on task and domain generalization \citep{muennighoff2023mtebmassivetextembedding}. As large language models (LLMs) have grown larger and more powerful, they have become integral in the ever-improving development of these general-purpose embedding models via techniques such as synthetic data generation and hard negative mining \citep{wang2024improvingtextembeddingslarge,lee2024gecko}, and by initializing embedding models with LLM weights \citep{ni2021largedualencodersgeneralizable}. As a result, much of the recent literature has centered around large models on the order of several billion parameters, such as NV-Embed \citep{lee2025nvembedimprovedtechniquestraining}, GritLM-7B \citep{muennighoff2025generativerepresentationalinstructiontuning}, and E5-Mistral \citep{wang2024improvingtextembeddingslarge}.

However, while the trend towards larger models has pushed the boundaries of performance, it has also highlighted a critical trade-off between capability and computational cost. State-of-the-art models are often too large and computationally expensive for many real-world applications, which often require low-latency, high-throughput inference. This is especially true for applications requiring on-device deployment, such as those involving sensitive data or offline access. This has led to a growing interest in developing lightweight models that can deliver strong performance without the need for extensive computational resources \citep{belcak2025smalllanguagemodelsfuture}, presenting a significant opportunity to further advance the capabilities of these smaller models.

\begin{figure}[!tb]
\centering
\includegraphics[width=0.495\textwidth]{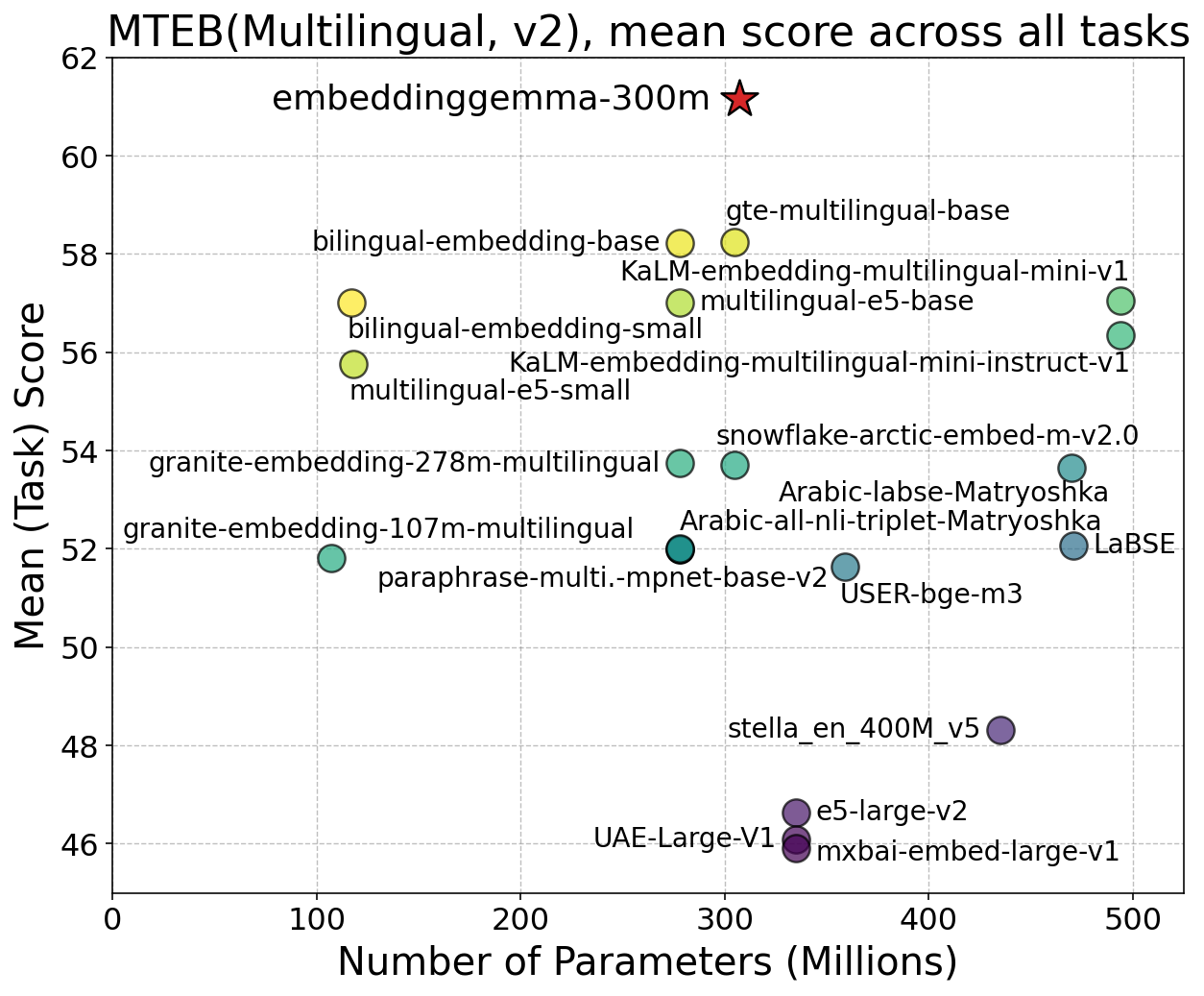}
\includegraphics[width=0.495\textwidth]{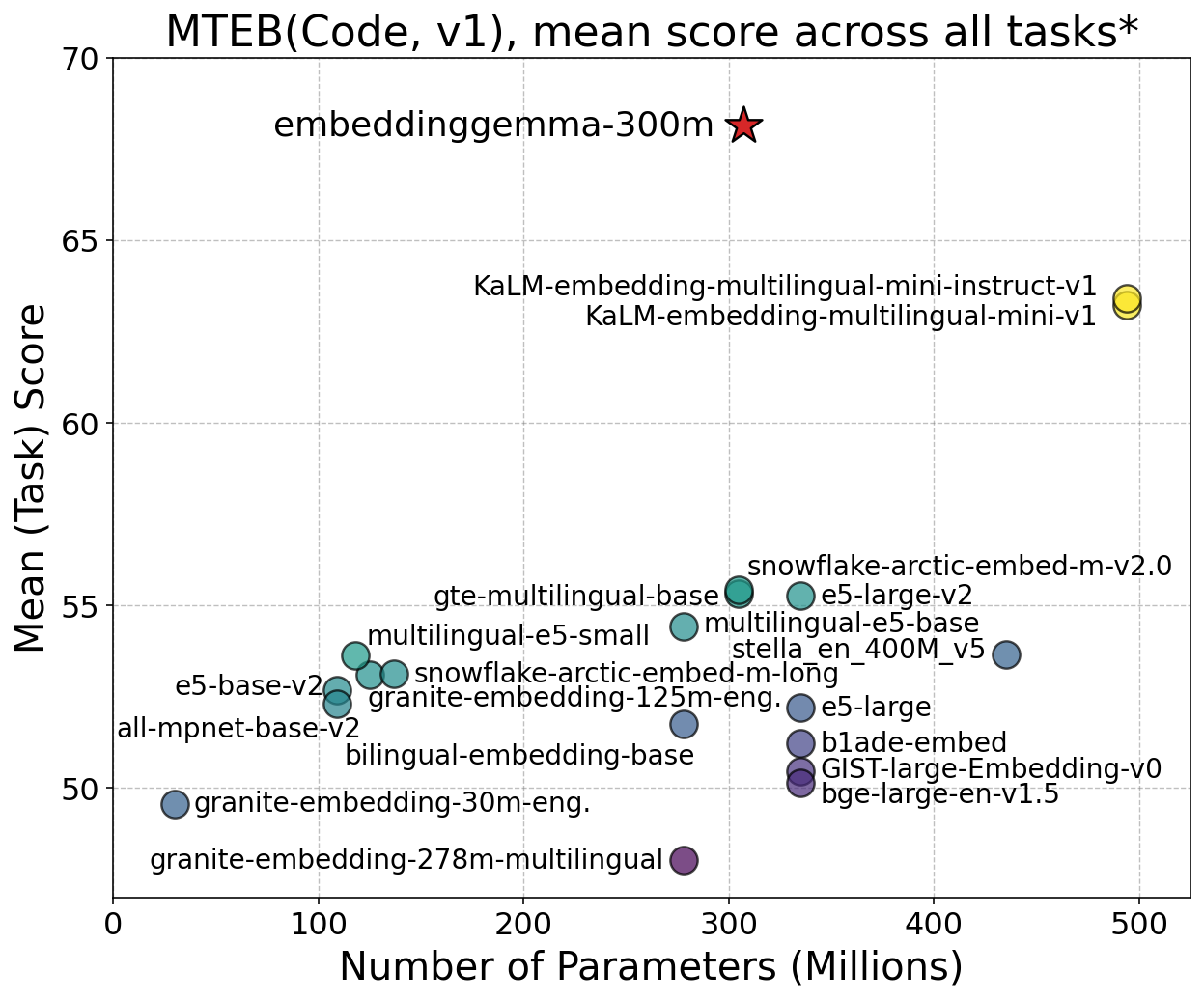}
\caption{Comparison of top 20 embedding models under 500M parameters across MTEB multilingual and code benchmarks. We exclude models trained on more than 25\% of the MTEB data to mitigate potential over‑fitting. $^*$: Average excludes COIRCodeSearchNetRetrieval, which was unavailable for most models. An analogous figure for MTEB(English, v2) is available in \Cref{sec:full_results_scatter}.
}
\label{fig:scatters}
\end{figure}

In this paper, we introduce EmbeddingGemma,\footnote{Our model is available at \url{https://ai.google.dev/gemma/docs/embeddinggemma}.} a novel, 308M parameter embedding model built on the architecture of the powerful Gemma 3 language model family \citep{gemma3}. In order to leverage richer encoder representations, we first adapt Gemma 3 into an encoder-decoder model using the UL2 \citep{tay2023ul2unifyinglanguagelearning} objective, following T5Gemma \citep{encdecgemma}. We then initialize EmbeddingGemma from the encoder of this encoder-decoder model. During training, we distill the state-of-the-art Gemini Embedding \citep{lee2025geminiembeddinggeneralizableembeddings} using embedding matching \citep{kim2023embeddistillgeometricknowledgedistillation}. We also include a ``spread-out'' regularizer to improve EmbeddingGemma's expressiveness and robustness. Building upon the success of Gecko and models that followed it \citep{lee2024gecko,lee2025geminiembeddinggeneralizableembeddings,zhang2025qwen3embeddingadvancingtext}, we include task prompts and a pre-finetuning stage to improve performance. Finally, we utilize model souping \citep{souping} to combine multiple finetuned checkpoints, opting to combine models trained on different finetuning mixtures rather than different hyperparameter configurations. This results in final embedding model with stronger, more generalizable representations.

To rigorously assess EmbeddingGemma's capabilities, we conduct extensive evaluations on multiple benchmarks. We mostly utilize the Massive Text Embedding Benchmark (MTEB) test suite \citep{enevoldsen2025mmteb, muennighoff2023mtebmassivetextembedding}; notably, its multilingual benchmark MTEB(Multilingual, v2) includes over 100 tasks spanning 250+ languages, 20 domains, and 9 task types such as classification, clustering, retrieval, and semantic similarity. As visualized in \Cref{fig:scatters}, EmbeddingGemma achieves state-of-the-art performance on MTEB(Multilingual, v2), MTEB(Code), and MTEB(English, v2) leaderboards for models under 500M parameters, ranking first across all aggregate metrics:\footnote{\url{https://huggingface.co/spaces/mteb/leaderboard}; September 23rd, 2025.} Borda count, mean over all task scores, and mean over all task type scores. Moreover, EmbeddingGemma represents a drastic improvement over the previous state-of-the-art: on MTEB(Multilingual, v2), for example, EmbeddingGemma ranks 8$^{\text{th}}$ overall, 17 places above the second-best sub-500M parameter model. EmbeddingGemma's performance is so strong it is comparable to models nearly double its size (\Cref{tab:main_table}). Impressively, EmbeddingGemma's lead persists even when truncating embeddings to as low as 128 dimensions or quantizing weights down to as low as 4-bit precision.

We also conduct detailed ablation studies to elucidate the factors contributing to the stellar performance of EmbeddingGemma. We find that though initializing model parameters from decoder-only LLMs indeed greatly improves performance \citep{lee2025geminiembeddinggeneralizableembeddings,zhang2025qwen3embeddingadvancingtext}, encoder-decoder models offer a stronger starting point. This is thanks to the encoder-decoder architecture's stronger contextual representations, which likely come from (i) the use of bidirectional attention and (ii) encoder parameters being able to specialize in input understanding \citep{encdecgemma,tay2023ul2unifyinglanguagelearning}. Counterintuitively, our experiments also show that simple pooling types tend to outperform attention pooling in embedding tasks despite using no learnable parameters. Our analysis also reveals that model souping works by varying finetuning mixtures instead of hyperparameters, perhaps in part due to our mixtures yielding models specialized in different areas. We also compare per-channel and per-block quantization-aware training and confirm they yield models robust to quantization.

\section{EmbeddingGemma}

\subsection{Architecture}
EmbeddingGemma is an encoder-only transformer model adapted from a pretrained 300M decoder-only Gemma 3 model. Specifically, we adapt the Gemma 3 model into an encoder-decoder model following the T5Gemma recipe \citep{encdecgemma}, and then initialize EmbeddingGemma from the encoder of this encoder-decoder model. This way, EmbeddingGemma is able to harness the extensive world knowledge of Gemma 3. Using an encoder-decoder adaptation further ensures the encoder is able to produce expressive representations from the start.

Specifically, given an input sequence $\mathbf{T}$ of $L$ tokens, we first apply $\mathcal{M}_n$, an $n$-layer transformer with bidirectional attention, and produce a sequence of token embeddings $\mathbf{T}_{\text{embed}} = \mathcal{M}_n(\mathbf{T}) \in \mathbb{R}^{L \times d_M}$, where $d_M$ is the model dimension used for the transformer's inner representations. We then employ a mean pooling $\mathcal{P}$, which averages the token embeddings along the sequence axis to generate a single embedding representing all the information in the input, producing $\mathbf{P}_{\text{embed}} = \mathcal{P}(\mathbf{T}_{\text{embed}}) \in \mathbb{R}^{d_M}$. Next, we apply a randomly initialized linear projection $g$ to upscale the embedding to an intermediate embedding dimension $d_U$, producing $\mathbf{E}_U = g(\mathbf{P}_{\text{embed}}) \in \mathbb{R}^{d_U}$. Finally, we apply another randomly initialized linear projection $f$ to scale the embedding to the target dimension $d$, resulting in $\mathbf{E} = f(\mathbf{E}_U) \in \mathbb{R}^d$. In EmbeddingGemma, we use $n=24, d_M=768, d_U=3072, d=768$.
\subsection{Training}
\paragraph{Input}
Each training example includes a query $q_i$, a positive passage $p_i^+$, and (optionally) a hard negative passage $p_i^-$. Each example also has prescribed task strings $t_q$ and $t_p$ for queries and passages respectively, describing the nature of the task. For instance, for retrieval we use ``\textit{task: search result | query:} \{content\}'' and ``\textit{title:} \{title | `none'\} \textit{| text:} \{content\}''. \\The query and passages are embedded as vectors in $\mathbb{R}^d$:
\begin{equation}
\mathbf{q}_i = f(g(\mathcal{P}(\mathcal{M}_n(t_q \oplus q_i)))), \quad \mathbf{p}_i^{\pm} = f(g(\mathcal{P}(\mathcal{M}_n(t_p \oplus p_i^{\pm})))).
\end{equation}
\paragraph{Objective}
EmbeddingGemma was trained using three different losses. The first is a noise-contrastive estimation (NCE) loss with in-batch negatives. Given a batch $\mathcal{B}$ of size $B := \vert \mathcal{B} \vert$, we define the contrastive loss as:
\begin{equation}
\mathcal{L_C} = \frac{1}{B} \sum_{i=1}^{B} \left[ -\log \frac{e^{\text{sim}(\mathbf{q}_i, \mathbf{p}_i^+)/\tau}}{w_i\, e^{\text{sim}(\mathbf{q}_i, \mathbf{p}_i^-)/\tau} + \sum_{j=1}^{B} \mathbb{1}_{\text{TN}}(i, j)\,e^{\text{sim}(\mathbf{q}_i, \mathbf{p}_j^+)/\tau}} \right]
\end{equation}
\vspace{0.1em}
where $\text{sim}(\mathbf{x}, \mathbf{y}) = \mathbf{x}^\top \mathbf{y} / \|\mathbf{x}\| \|\mathbf{y}\|$ is cosine similarity, and $\mathbb{1}_{\text{TN}}$ masks out false negatives from duplicates:
\begin{equation}
\mathbb{1}_{\text{TN}}(i, j) =
\begin{cases}
    0 & \text{if } q_i = q_j \text{ or } p_i^+ = p_j^+, \\
    1 & \text{otherwise.}
\end{cases}
\end{equation}
\vspace{0.1em}
The hardness weight $w_i$ represents how challenging a (query, hard negative passage) pair is for the model to differentiate between, forcing it to learn more discriminative representations \citep{lan2025llavelargelanguagevision}. This is defined as $w_i = \exp({\alpha\,\text{sg}(\text{sim}(\mathbf{q}_i, \mathbf{p}_i^-)))}$, where $\text{sg}(\cdot)$ is the stop-gradient operator, and $\alpha$ is a hyperparameter that controls the strength of the weighting, experimentally set to $5.0$. The stop-gradient ensures we are weighing based on the current difficulty and not differentiating through the weight factor itself.

The second loss is based on the global orthogonal regularizer (GOR) introduced in \cite{spreadout}. This loss encourages EmbeddingGemma to produce embeddings that are spread out over the embedding space, to fully utilize the expressive power of the embedding space. This also intends to ensure that a) the model is robust to quantization (especially embedding quantization), and that b) the embeddings produced by the model can be retrieved efficiently in vector databases using approximate nearest neighbor (ANN) algorithms. We define the spread-out loss as:
\vspace{0.1em}
\begin{align}
\mathcal{L_S} &= \frac{1}{B\,(B-1)} \sum_{i,j \in \mathcal{B}\,:\,i \ne j} (\mathbf{q}_{i}^\top \mathbf{q}_{j})^2 +\, \frac{1}{B\,(B-1)} \sum_{i,j \in \mathcal{B}\,:\,i \ne j} (\mathbf{p}_{i}^{+\top} \mathbf{p}_{j}^+)^2 .
\end{align}
\vspace{0.1em}
The idea is to make the embeddings of a random pair of inputs have similar statistical properties (mean and second moment) as two points independently and uniformly sampled from the unit sphere. We use only the ``second moment'' term of the regularizer's original definition, as we find this also sufficiently pushes the ``mean'' term towards its target value.

The third and final is an embedding matching loss based on \cite{kim2023embeddistillgeometricknowledgedistillation}. In contrast to previous distillation research which has relied only on the teacher's query-document relevance scores as a signal \citep{moreira2025nvretrieverimprovingtextembedding,santhanam2022colbertv2effectiveefficientretrieval}, this loss directly aligns EmbeddingGemma's embedding space with that of a teacher model, allowing it to more effectively learn from the larger, more powerful Gemini Embedding model. Note that unlike \cite{kim2023embeddistillgeometricknowledgedistillation}, we apply embedding matching not only to queries and passages but also to hard negative passages, as we found this substantially improves performance. Intuitively, this serves the same purpose as using hard negative passages in NCE loss, as the model learns how the teacher discriminates between the queries and its corresponding hard negatives. We weigh these individual losses uniformly, letting
\vspace{0.1em}
\begin{align}
\mathcal{L_D} &= \mathcal{L^{\text{Q}}_D} + \mathcal{L^{\text{P}^+}_D} + \mathcal{L^{\text{P}^-}_D}.
\end{align}

We adapt the contrastive and spread-out losses using MRL \citep{kusupati2024matryoshkarepresentationlearning}, which splits each loss into $k$ separate losses applied to $k$ overlapping sub-dimensions of the embedding. We consider each of these individual losses equally, simply adding their values during training without any special weights. EmbeddingGemma provides $d=768$ dimensional embeddings, additionally supporting $512$, $256$, and $128$ dimensional embeddings via MRL.

\subsection{Recipe}
\label{sec:recipe}

We train EmbeddingGemma on a mixture of embedding tasks spanning various languages and task types. In total, including encoder-decoder training, EmbeddingGemma sees approximately 2.1T tokens, of which 314B are seen during pre-finetuning and 20B during finetuning.

\paragraph{Encoder-Decoder Training}
We begin by adapting the decoder-only Gemma 3 model to an encoder-decoder model to get a strong encoder with improved contextual representations. Following T5Gemma \citep{encdecgemma}, we initialize the encoder-decoder with the Gemma 3 decoder-only checkpoint and further pretrain it on the Gemma 3 pretraining data with UL2 \citep{tay2023ul2unifyinglanguagelearning}. We utilize the pretrained encoder as the backbone for our embedding model, providing EmbeddingGemma with a strong initialization from which it inherits extensive world knowledge, such as the 100+ languages used to train Gemma 3.

\paragraph{Pre-finetuning}
We then train the embedding model itself, starting with a pre-finetuning stage. Here we train on large-scale unsupervised data, as has been previously shown to improve performance \citep{DBLP:journals/corr/abs-2112-09118,gao2022simcsesimplecontrastivelearning}. We use only (query, target) pairs as input, as the training mixture is large and noisy, so mining high-quality hard negatives is challenging. We opt for a larger batch size in order to stabilize the gradient, and because this improves performance by effectively providing more (in-batch) negatives for each input example.

The goal of this stage is to build the model's generalization capabilities by training for a large number of steps over a diverse training set. As such, our pre-finetuning mixture spans both various, evenly weighted task types---question answering, sentence similarity, code retrieval, and web search tasks---and various languages (natural and programming). As part of this, we leverage a corpus containing billions of title and body text pairs crawled from websites, much like in previous work \citep{lee2024gecko,wang2024textembeddingsweaklysupervisedcontrastive,neelakantan2022textcodeembeddingscontrastive}.

\paragraph{Finetuning}
Next, we finetune the model using a smaller but higher-quality mixture of task-specific datasets. Here, we opt for a smaller batch size, and do utilize hard negatives. Note that both here and in pre-finetuning, each batch only contains examples from a given dataset, as in-batch negatives are more difficult to discern, and thus more valuable for contrastive learning, coming from the same task.

As in Gemini Embedding \citep{lee2025geminiembeddinggeneralizableembeddings}, we utilize a set of three different groups of tasks aimed at task diversity, language diversity, and coding capability, respectively. This includes a subset of the academic datasets used by Gecko \citep{lee2024gecko} and the synthetic datasets used by Gemini Embedding. However, instead of directly determining the task mixture rates based on a fine-grained grid search, we use the mixture resulting from this procedure as a seed for Bayesian optimization, alongside 10 other random mixtures sampled from a Dirichlet distribution. This yields multiple mixtures which individually result in stronger performance than the seeds. Moreover, thanks to a balance of explore-exploit objectives, these mixtures specialize in different domains, creating experts which synergize during souping.

\paragraph{Model Souping}
\label{sec:model_souping}
Finally, we combine models from our finetuning runs by averaging their parameters, to improve our final model's quality and robustness. We assessed various methods including merging checkpoints from various steps of a single training run \citep{same_run_souping} and from multiple training runs using different hyperparameters \citep{souping}. We find that mixtures obtained from Bayesian optimization work well together as souping ingredients: our final model is the result of an unweighted average of checkpoints from finetuning runs with each of the mixtures obtained with Bayesian optimization.

\begin{table}[t]
\centering
\resizebox{0.75\columnwidth}{!}{%
\begin{tabular}{l|cc|cc|c}
\toprule
 & \multicolumn{2}{c}{MTEB(Multi, v2)} & \multicolumn{2}{c}{MTEB(Eng, v2)} & \multicolumn{1}{c}{MTEB(Code)}\\
\cmidrule(lr){2-3} \cmidrule(lr){4-5} \cmidrule(lr){6-6}
Precision & Mean(Task) & Mean(Type) & Mean(Task) & Mean(Type) & Mean(Task)\\
\midrule
bf16 & 61.15 & 54.31 & 69.67 & 65.11 & 68.76 \\
int8$^*$ & 60.93 & 53.95 & 69.49 & 64.84 & 68.70\\
Mixed$^\dagger$ & 60.69 & 53.82 & 69.32 & 64.82 & 68.03\\
int4$^*$ & 60.62 & 53.61 & 69.31 & 64.65 & 67.99 \\
\bottomrule
\end{tabular}
}
\caption{Performance of raw and quantized EmbeddingGemma checkpoints on MTEB benchmarks.\\$^*$: Per-block quantization. $^\dagger$: Per-channel quantization with int4 for embedding, feedforward, and projection layers, and int8 for attention}
\label{tab:qat_quality_table}
\vspace{-0.2em}
\end{table}

\vspace{-0.75em}
\paragraph{Quantization-Aware Training}
We additionally provide quantized versions of EmbeddingGemma in standard quantization configurations. We provide checkpoints with three types of weight representations: int4 per-block, int8 per-block, and mixed-precision per-channel. We obtain these variants by applying quantization-aware training \citep{8578384} during the model finetuning stage, to minimize quality degradation for quantized checkpoints. \Cref{tab:qat_quality_table} shows the quality is comparable between our quantized checkpoints and our raw checkpoint (evaluated at half-precision bf16 weights).
\vspace{-1.1em}
\section{Ablation Studies}
\label{sec:ablations}
We analyze some key design choices to better understand how they contributed to EmbeddingGemma's ability to produce high-quality embeddings across tasks and languages despite its small size. Models for ablations are finetuned on only one mixture and thus exclude model souping, unless indicated otherwise. We evaluate the models on MTEB(Multilingual, v2), as it covers various languages, domains, and task types.

\begin{table}[t]
\centering
\resizebox{1\columnwidth}{!}{%
\begin{tabular}{lcc|ccccccccc}
\toprule
& \multirow{2}{*}{\parbox{1cm}{\centering{\textbf{Mean}\\\textbf{(Task)}}}} & \multirow{2}{*}{\parbox{1cm}{\centering \textbf{Mean}\\\textbf{(Type)}}} & \multirow{2}{*}{\parbox{1.2cm}{\centering Bitext\\Mining}} & & & \multirow{2}{*}{\parbox{1.2cm}{\centering Inst.\\Retrieval}} & \multirow{2}{*}{\parbox{1.2cm}{\centering Multi.\\Class.}} & \multirow{2}{*}{\parbox{1.2cm}{\centering Pair\\Class.}} \\
Initialization &  & & & {Class.} & {Clus.} & & & & {Rerank.} & {Retrieval} & {STS} \\
\midrule
Encoder-Decoder & \textcolor{blue}{\textbf{60.4}} & \textcolor{blue}{\textbf{53.6}} & \textcolor{blue}{\textbf{63.5}} & \textcolor{blue}{\textbf{60.4}} & 50.3 & \textcolor{blue}{\textbf{4.3}} & \textcolor{blue}{\textbf{24.6}} & \textcolor{blue}{\textbf{81.1}} & \textcolor{blue}{\textbf{63.1}} & \textcolor{blue}{\textbf{60.2}} & \textcolor{blue}{\textbf{74.5}} \\
\midrule
Decoder-only & 59.7 & 52.6 & 63.2 & 60.1 & \textbf{50.5} & 0.8 & 24.2 & 80.4 & 61.6 & 58.3 & 73.9 \\
Random & 45.2 & 39.2 & 26.8 & 48.8 & 41.5 & -0.2 & 16.4 & 72.5 & 49.1 & 35.5 & 62.1 \\
\bottomrule
\end{tabular}
}
\caption{Results using different initialization strategies.}
\label{tab:init_ablation}
\vspace{-0.4em}
\end{table}

\vspace{-0.5em}
\subsection{Initialization Strategy}
\Cref{tab:init_ablation} compares model performance when initialized with a decoder-only Gemma 3 model and with an encoder-decoder based on that Gemma 3 model, trained as described in \Cref{sec:recipe}. Encoder-decoder initialization outperforms decoder-only initialization across various task types. This suggests the encoder-decoder is able to produce more expressive representations, as put forward in the literature. As a baseline, we include the performance of a model initialized with random weights, demonstrating both initialization strategies dramatically improve model performance.

\begin{table}[t]
\centering
\resizebox{1\columnwidth}{!}{%
\begin{tabular}{lcc|ccccccccc}
\toprule
& \multirow{2}{*}{\parbox{1cm}{\centering{\textbf{Mean}\\\textbf{(Task)}}}} & \multirow{2}{*}{\parbox{1cm}{\centering \textbf{Mean}\\\textbf{(Type)}}} & \multirow{2}{*}{\parbox{1.2cm}{\centering Bitext\\Mining}} & & & \multirow{2}{*}{\parbox{1.2cm}{\centering Inst.\\Retrieval}} & \multirow{2}{*}{\parbox{1.2cm}{\centering Multi.\\Class.}} & \multirow{2}{*}{\parbox{1.2cm}{\centering Pair\\Class.}} \\
Pooling &  & & & {Class.} & {Clus.} & & & & {Rerank.} & {Retrieval} & {STS} \\
\midrule
Mean & \textcolor{blue}{\textbf{60.4}} & \textcolor{blue}{\textbf{53.6}} & \textcolor{blue}{\textbf{63.5}} & \textcolor{blue}{\textbf{60.4}} & \textcolor{blue}{\textbf{50.3}} & \textcolor{blue}{\textbf{4.3}} & \textcolor{blue}{\textbf{24.6}} & 81.1 & \textcolor{blue}{\textbf{63.1}} & 60.2 & \textcolor{blue}{\textbf{74.5}} \\
\midrule
Last Token & 59.7 & 52.6 & 63.1 & 60.3 & 49.9 & 1.9 & 23.1 & 80.8 & 62.4 & 57.8 & 73.8 \\
First Token & 59.9 & 52.9 & \textbf{63.5} & 60.3 & 49.7 & 2.2 & 23.8 & \textbf{81.4} & 62.6 & 58.6 & 73.8 \\
Attention & 60.2 & 53.1 & \textbf{63.5} & 60.1 & 49.6 & 1.5 & 23.4 & 80.9 & 62.7 & \textbf{61.7} & 74.1 \\
\bottomrule
\end{tabular}
}
\caption{Results using different types of poolers.}
\label{tab:pool_ablation}
\end{table}

\subsection{Pooling Types}
Poolers aggregate the transformer's token representations into a single vector. In \textit{mean pooling}, we compute the average of all the token representations. In \textit{first-token pooling}, we simply take the representation of the first token and use it directly. In \textit{last-token pooling}, we do the same, but with the last token of the sequence. In \textit{attention pooling}, we utilize an attention mechanism to weigh and aggregate the token representations. Our pooled representations can be expressed as $\mathbf{P}_{\text{embed}} = \text{softmax}(\frac{QK^\top}{\sqrt{d_M}})V \in \mathbb{R}^{1 \times d_M}$, where the input is passed through the key and value matrices, $K,V \in \mathbb{R}^{L \times d_M}$, and the query matrix $Q \in \mathbb{R}^{1 \times d_M}$ is a learnable parameter. We specifically employ multi-head attention with four attention heads.

The results in \Cref{tab:pool_ablation} indicate that mean pooling yields the best performance, despite attention pooling offering a large amount of additional learnable parameters. This is consistent with the results from \cite{suganthan2025adaptingdecoderbasedlanguagemodels}, which found that simple pooling strategies outperformed attention pooling in encoder-only models for classification and regression tasks, suggesting that these results further extend to embedding tasks such as clustering.

\begin{table}[t]
\centering
\resizebox{1\columnwidth}{!}{%
\begin{tabular}{lcc|ccccccccc}
\toprule
& \multirow{2}{*}{\parbox{1cm}{\centering{\textbf{Mean}\\\textbf{(Task)}}}} & \multirow{2}{*}{\parbox{1cm}{\centering \textbf{Mean}\\\textbf{(Type)}}} & \multirow{2}{*}{\parbox{1.2cm}{\centering Bitext\\Mining}} & & & \multirow{2}{*}{\parbox{1.2cm}{\centering Inst.\\Retrieval}} & \multirow{2}{*}{\parbox{1.2cm}{\centering Multi.\\Class.}} & \multirow{2}{*}{\parbox{1.2cm}{\centering Pair\\Class.}} \\
Pooling &  & & & {Class.} & {Clus.} & & & & {Rerank.} & {Retrieval} & {STS} \\
\midrule
Souped & \textcolor{blue}{\textbf{61.2}} & \textcolor{blue}{\textbf{54.3}} & \textcolor{blue}{\textbf{64.4}} & \textcolor{blue}{\textbf{60.9}} & \textcolor{blue}{\textbf{51.2}} & \textcolor{blue}{\textbf{5.6}} & \textcolor{blue}{\textbf{24.8}} & \textcolor{blue}{\textbf{81.4}} & \textcolor{blue}{\textbf{63.3}} & \textcolor{blue}{\textbf{62.5}} & \textcolor{blue}{\textbf{74.7}} \\
\midrule
Mix 1 & \underline{60.4} & 53.4 & 63.5 & \underline{60.4} & \underline{50.6} & 4.1 & 23.1 & 81.1 & 63.0 & \underline{61.0} & 73.9 \\
Mix 2 & \underline{60.4} & \underline{53.6} & 63.5 & \underline{60.4} & 50.3 & \underline{4.3} & \underline{24.6} & 81.1 & \underline{63.1} & 60.2 & \underline{74.5} \\
Mix 3 & 60.1 & 53.3 & \underline{63.9} & \underline{60.4} & 50.5 & 3.6 & 24.3 & \underline{\textbf{81.4}} & 62.8 & 58.2 & \underline{74.5} \\
\bottomrule
\end{tabular}
}
\caption{Results using different training mixtures. Best result between the three mixtures is underlined.}
\label{tab:soup_ablation}
\end{table}

\subsection{Model Souping}
Here, we train models on different model soup ingredients and compare their performance to our final souped checkpoint: \Cref{tab:soup_ablation}. The model soup not only improves on overall performance, but even outperforms the ingredients in each task type. This indicates that model souping works not only on runs with varied hyperparameter configurations, but also on runs with different finetuning mixtures altogether. Note also that each mixture yields an expert in different task types.

\section{Evaluation}

We evaluate EmbeddingGemma on a comprehensive set of benchmarks covering diverse task types, domains, languages, and language pairs. Specifically, we report numbers for the Massive Text Embedding Benchmark \citep{enevoldsen2025mmteb, muennighoff2023mtebmassivetextembedding}, XOR-Retrieve \citep{asai2021xor}, and XTREME-UP \citep{ruder2023xtreme}.
\newcommand{\rotheader}[1]{\rotatebox{65}{\makecell[l]{#1}}}
\newcommand{\mcrot}[4]{\multicolumn{#1}{#2}{\rlap{\rotatebox{#3}{#4}~}}}
\newcommand{\rotangle}{45}

\begin{table}[!t]
\centering
\resizebox{0.95\columnwidth}{!}{%
\begin{tabular}{llcc|ccccccc|cc}
\Xcline{1-9}{0.08em} \Xcline{11-13}{0.08em}
\rule{0pt}{3ex} & & \multicolumn{2}{l}{Open Models} & & & & & & & \multicolumn{3}{l}{Commercial API Models} \\
\cline{1-9} \cline{11-13}
\rule{0pt}{3ex} 
& & \mcrot{1}{l}{\rotangle}{\shortstack[l]{Embedding\\Gemma}} 
    & \mcrot{1}{l}{\rotangle}{\shortstack[l]{Gecko\\Embedding$^\dagger$}} 
    & \mcrot{1}{l}{\rotangle}{\shortstack[l]{GTE\\Multilingual Base}} 
    & \mcrot{1}{l}{\rotangle}{\shortstack[l]{mE5 Large\\Instruct}} 
    & \mcrot{1}{l}{\rotangle}{BGE-M3} 
    & \mcrot{1}{l}{\rotangle}{\shortstack[l]{Jina\\Embeddings v3}}
    & \mcrot{1}{l}{\rotangle}{\shortstack[l]{Qwen3\\Embedding 0.6B}} 
    &
    & \mcrot{1}{l}{\rotangle}{\shortstack[l]{Gemini\\Embedding}} 
    & \mcrot{1}{l}{\rotangle}{\shortstack[l]{Cohere Embed\\Multilingual v3}}
    & \mcrot{1}{l}{\rotangle}{\shortstack[l]{text-embedding-\\3-large}} \\
\cmidrule{1-9} \cmidrule{11-13}
\textbf{Parameters} & & 308M & 278M & 305M & 560M & 568M & 572M & 595M & & -- & -- & -- \\
\cmidrule{1-9} \cmidrule{11-13}
\textbf{Memory Usage (MB)} & & 578 & -- & 582 & 1068 & 2167 & 1092 & 2272 & & -- & -- & -- \\
\cmidrule{1-9} \cmidrule{11-13}
\multirow{2}{3.5cm}{\textbf{MTEB(Multi, v2)} \footnotesize{\citep{enevoldsen2025mmteb}}} & Mean (Task) & 61.15 & 53.47 & 58.24 & 63.22 & 59.56 & 58.37 & \textbf{64.34} & & 68.37 & 61.12 & 58.93 \\
& Mean (Type) & 54.31 & 46.23 & 51.44 & 55.08 & 52.18 & 50.66 & \textbf{56.01} & & 59.59 & 53.23 & 51.41\\
\cmidrule{2-9} \cmidrule{11-13}
& - Bitext Mining & 64.40 & 43.57 & 71.79 & \textbf{80.13} & 79.11 & 65.25 & 72.23 & & 79.28 & 70.50 & 62.17 \\
& - Classification & 60.90 & 55.99 & 57.17 & 64.94 & 60.35 & 58.77 & \textbf{66.83} & & 71.82 & 62.95 & 60.27 \\
& - Clustering & 51.17 & 43.83 & 44.33 & 50.75 & 40.88 & 45.65 & \textbf{52.33} & & 54.59 & 46.89 & 46.89\\
& - Inst. Retrieval & \textcolor{blue}{\textbf{5.61}} & 0.42 & -0.74 & -0.40 & -3.11 & -1.34 & 5.09 & & 5.18 & -1.89 & -2.68 \\
& - Multilabel Class. & \textcolor{blue}{\textbf{24.82}} & 17.00 & 19.82 & 22.91 & 20.10 & 18.38 & 24.59 & & 29.16 & 22.74 & 22.03\\
& - Pair Class. & \textcolor{blue}{\textbf{81.40}} & 75.61 & 80.49 & 80.86 & 80.76 & 79.27 & 80.83 & & 83.63 & 79.88 & 79.17 \\
& - Reranking & \textcolor{blue}{\textbf{63.25}} & 57.07 & 60.72 & 62.61 & 62.79 & 57.09 & 61.41 & & 65.58 & 64.07 & 63.89\\
& - Retrieval & 62.49 & 53.77 & 56.50 & 57.12 & 54.60 & 55.76 & \textbf{64.65} & & 67.71 & 59.16 & 59.27\\
& - STS & 74.73 & 68.83 & 72.75 & 76.81 & 74.12 & \textbf{77.13} & 76.17 & & 79.40 & 74.80 & 71.68 \\
\cmidrule{1-9} \cmidrule{11-13}
\multirow{2}{3.5cm}{\textbf{MTEB(Eng, v2)} \footnotesize{\citep{enevoldsen2025mmteb}}} & Mean (Task) & 69.67 & 66.38 & -- & 65.53 & -- & -- & \textbf{70.70} & & 73.30 & 66.01 & 66.43 \\
& Mean (Type) & \textcolor{blue}{\textbf{65.11}} & 62.33 & -- & 61.21 & -- & -- & 64.88 & & 67.67 & 61.43 & 62.15  \\
\cmidrule{1-9} \cmidrule{11-13}
\multirow{2}{3.5cm}{\textbf{MTEB(Code)}$^*$ \footnotesize{\citep{enevoldsen2025mmteb}}} & & \multirow{2}{*}{68.14} & \multirow{2}{*}{--} & \multirow{2}{*}{55.33} & \multirow{2}{*}{60.55} & \multirow{2}{*}{--} & \multirow{2}{*}{58.03} & \multirow{2}{*}{\textbf{74.57}} & & \multirow{2}{*}{75.54} & \multirow{2}{*}{57.26} & \multirow{2}{*}{65.73} \\
& & & & & & & & & & & & \\
\cmidrule{1-9} \cmidrule{11-13}
\multirow{2}{3cm}{\textbf{XOR-Retrieve} \footnotesize{\citep{asai2021xor}}} & & \multirow{2}{*}{\textcolor{blue}{\textbf{84.14}}} & \multirow{2}{*}{41.24} & \multirow{2}{*}{--} & \multirow{2}{*}{--} & \multirow{2}{*}{--} & \multirow{2}{*}{--} & \multirow{2}{*}{--} & & \multirow{2}{*}{90.42} & \multirow{2}{*}{--} & \multirow{2}{*}{68.76} \\
& & & & & & & & & & & & \\
\cmidrule{1-9} \cmidrule{11-13}
\multirow{2}{3.5cm}{\textbf{XTREME-UP}\\ \footnotesize{\citep{ruder2023xtreme}}} & & \multirow{2}{*}{\textcolor{blue}{\textbf{47.72}}} & \multirow{2}{*}{7.61} & \multirow{2}{*}{19.05} & \multirow{2}{*}{18.68} & \multirow{2}{*}{26.91} & \multirow{2}{*}{8.54} & \multirow{2}{*}{6.64} & & \multirow{2}{*}{64.33} & \multirow{2}{*}{--} & \multirow{2}{*}{18.80} \\
& & & & & & & & & & & & \\
\Xcline{1-9}{0.08em} \Xcline{11-13}{0.08em}
\end{tabular}
}
\vspace{0.2em}
\caption{Comparison of popular embedding models on the Massive Text Embedding Benchmark: MTEB(Multilingual, v2), MTEB(English, v2), and MTEB(Code). We also show results on XOR-Retrieve and XTREME-UP, reporting Recall@5kt and MRR@10 respectively. $^*$: Averaged over all code tasks excluding COIRCodeSearchNetRetrieval, which was unavailable for most models. $^\dagger$: For Gecko Embedding \citep{lee2024gecko}, we evaluate a smaller version of the model discussed in the paper.
}
\label{tab:main_table}
\vspace{0.8em}
\end{table}

\subsection{Tasks}
MTEB consists of a massive collection of individual, quality-controlled evaluation tasks. Together, the subsets we consider cover 250+ natural languages and 14+ programming languages, 20 domains (e.g. legal, medical, news, programming, web), and 10 task types: bitext mining, classification, clustering, instruction retrieval, multilabel classification, pair classification, reranking, retrieval, semantic text similarity, and summarization. Our evaluations include 162 individual tasks, consisting of 131 tasks for MTEB(Multilingual, v2), 41 tasks for MTEB(English, v2), and 12 code retrieval tasks for MTEB(Code). XOR-Retrieve and XTREME-UP provide cross-lingual retrieval evaluations, with XOR-Retrieve pairing English passages with retrieval queries in 7 different languages and XTREME-UP pairing English passages with queries in 20 underrepresented Indo-European languages.

\subsection{Methods}
We compare our model to the most popular general-purpose embedding models in two categories: open embedding models with fewer than a billion parameters and models available through commercial APIs. For MTEB benchmarks, we additionally compare against the top models with fewer than 500M parameters in the respective MTEB leaderboards. We exclude models trained on more than 25\% of the MTEB data to mitigate potential overfitting.
\vspace{0.3em}

We run our evaluations with half-precision weights (bfloat16), using the prompt instructions detailed in the model card.\footnote{\url{https://ai.google.dev/gemma/docs/embeddinggemma/model_card\#prompt-instructions}} We use a context length of 512 tokens for most evaluation tasks, increasing to 1024 or 2048 for tasks such as LongEmbed Passkey Retrieval \citep{longembed}, which require longer input sequences.

\vspace{-0.2em}
\subsection{Results}
\paragraph{Overall Performance} In \Cref{tab:main_table}, we present EmbeddingGemma's performance across MTEB benchmarks, XOR-Retrieve, and XTREME-UP, comparing it to popular general-purpose embedding models. Below, we also provide more thorough comparisons of MTEB benchmark performance, reporting performance for other top models as well as for EmbeddingGemma's lower-dimensional embeddings.

EmbeddingGemma pushes the limit for quality across several benchmarks. \textbf{EmbeddingGemma achieves the \#1 rank and highest overall performance on the MTEB multilingual, English, and code leaderboards across models under 500M parameters}, with a significant lead over all previous top performing models on each of the metrics summarizing aggregate performance across tasks (Task Mean, Task Type Mean, and Borda rank). This gap persists when using fewer embedding dimensions: \textbf{Even with 128-dimensional embeddings, EmbeddingGemma achieves the highest scores for Task Mean and Task Type Mean}. EmbeddingGemma also achieves excellent scores on XOR-Retrieve and XTREME-UP; in sum, these results demonstrate it is a state-of-the-art general-purpose model with proven capabilities spanning cross-lingual, multilingual, English, and code tasks.

This performance advantage is not limited to small models: \textbf{EmbeddingGemma is competitive with larger models}, providing performance comparable to state-of-the-art models nearly double its size and with larger embedding dimensions; ranking \#3, \#2, and \#2 across models under 1B parameters on the MTEB multilingual, English, and code leaderboards respectively; and outperforming commercial API models, with the notable exception of Gemini Embedding.

\begin{table}[!t]
\centering
\resizebox{\columnwidth}{!}{%
\begin{tabular}{lccccc|ccccccccc}
\toprule
& \multirow{2}{*}{\parbox{1cm}{\centering Memory\\(MB)}} & & & \multirow{2}{*}{\parbox{1cm}{\centering{\textbf{Mean}\\\textbf{(Task)}}}} & \multirow{2}{*}{\parbox{1cm}{\centering \textbf{Mean}\\\textbf{(Type)}}} & \multirow{2}{*}{\parbox{1.2cm}{\centering Bitext\\Mining}} & & & \multirow{2}{*}{\parbox{1.2cm}{\centering Inst.\\Retrieval}} & \multirow{2}{*}{\parbox{1.2cm}{\centering Multi.\\Class.}} & \multirow{2}{*}{\parbox{1.2cm}{\centering Pair\\Class.}} \\
Model Name & & Parameters & \textbf{Rank} &  &  & & {Class.} & {Clus.} & & & & {Rerank.} & {Retrieval} & {STS} \\
\midrule
{EmbeddingGemma (768d)} & 578 & 308M & \textcolor{blue}{\textbf{8}} & \textcolor{blue}{\textbf{61.2}} & \textcolor{blue}{\textbf{54.3}} & 64.4 & \textcolor{blue}{\textbf{60.9}} & \textcolor{blue}{\textbf{51.2}} & \textcolor{blue}{\textbf{5.6}} & \textcolor{blue}{\textbf{24.8}} & \textcolor{blue}{\textbf{81.4}} & \textcolor{blue}{\textbf{63.3}} & \textcolor{blue}{\textbf{62.5}} & 74.7 \\
\hphantom{EmbeddingGemma} (512d) & & & & 60.7 & 53.9 & 64.0 & 60.3 & 50.9 & 4.8 & 24.5 & 81.5 & 62.7 & 61.5 & 74.8 \\
\hphantom{EmbeddingGemma} (256d) & & & & 59.7 & 53.0 & 62.4 & 59.3 & 50.5 & 4.4 & 23.9 & 81.5 & 61.8 & 58.8 & 74.6 \\
\hphantom{EmbeddingGemma} (128d) & & & & 58.2 & 51.8 & 59.3 & 57.8 & 50.6 & 3.8 & 23.1 & 81.3 & 60.8 & 55.3 & 73.8 \\
\midrule

KaLM mini-v1 & 1885 & 494M & 25 & 57.0 & 50.0 & 64.8 & 57.6 & 45.6 & -1.5 & 20.7 & 77.7 & 60.6 & 54.2 & 70.8\\
gte-multilingual-base & 582 & 305M & 26 & 58.2 & 51.4 & \textbf{71.8} & 57.2 & 44.3 & -0.7 & 19.8 & 80.5 & 60.7 & 56.5 & 72.9\\
bilingual-embedding-base & 1061 & 278M & 28 & 58.2 & 50.6 & 70.0 & 59.4 & 44.7 & -3.8 & 20.7 & 77.4 & 59.4 & 53.0 & \textbf{74.8}\\
KaLM mini-instruct-v1 & 1885 & 494M & 29 & 56.3 & 49.3 & 64.2 & 57.4 & 45.9 & -2.6 & 21.6 & 77.5 & 58.2 & 50.8 & 70.5\\
bilingual-embedding-small & 449 & 117M & 34 & 57.0 & 49.7 & 69.5 & 58.1 & 43.8 & -3.4 & 19.2 & 77.1 & 59.3 & 49.5 & 74.1\\
multilingual-e5-base & 1061 & 278M & 36 & 57.0 & 49.8 & 69.4 & 58.2 & 41.8 & -2.7 & 20.2 & 77.2 & 60.2 & 52.7 & 71.4\\
multilingual-e5-small & 449 & 118M & 39 & 55.8 & 49.0 & 69.5 & 56.2 & 40.8 & -2.4 & 19.0 & 76.9 & 60.4 & 49.3 & 71.7\\
snowflake-arctic-embed-m-v2.0 & 1165 & 305M & 41 & 53.7 & 46.9 & 53.7 & 54.4 & 42.2 & -3.3 & 17.0 & 74.9 & 61.7 & 54.8 & 66.6\\
USER-bge-m3 & 1370 & 359M & 42 & 51.6 & 45.3 & 63.4 & 53.2 & 40.0 & -2.7 & 18.6 & 80.6 & 51.5 & 43.3 & 59.5 \\
Arabic-labse-Matryoshka & 1796 & 470M & 44 & 53.6 & 47.2 & 70.3 & 55.1 & 39.6 & -1.8 & 19.9 & 80.8 & 51.6 & 37.0 & 72.1\\
granite-embedding-278m-multi. & 530 & 278M & 45 & 53.7 & 47.2 & 58.5 & 54.1 & 41.4 & -1.8 & 17.7 & 75.7 & 59.9 & 52.2 & 67.3\\
\bottomrule
\end{tabular}
}
\caption{Performance of top leaderboard models under 500M parameters on MTEB(Multilingual, v2). Rank is computed using Borda count against all models, regardless of parameter count.}
\label{tab:leaderboard_multilingual}
\vspace{-0.2em}
\end{table}
\vspace{-0.2em}
\paragraph{MTEB(Multilingual, v2)}
We compare EmbeddingGemma to other top-ranked models from the MTEB multilingual benchmark in \Cref{tab:leaderboard_multilingual}. Besides achieving the highest values for the aggregate scores, it leads in nearly all task types. Moreover, EmbeddingGemma offers drastic improvements in each of the task types it excels at, even going so far as to compete with larger models. For example, it outperforms Qwen3 Embedding 0.6B, the top-ranked model with fewer than 1B parameters, in instruction retrieval, multilabel classification, pair classification, and reranking (\Cref{tab:main_table}).

\begin{table}[!t]
\centering
\resizebox{\columnwidth}{!}{%
\begin{tabular}{lccccc|ccccccc}
\toprule
& \multirow{2}{*}{\parbox{1cm}{\centering Memory\\(MB)}} & & & \multirow{2}{*}{\parbox{1cm}{\centering{\textbf{Mean}\\\textbf{(Task)}}}} & \multirow{2}{*}{\parbox{1cm}{\centering \textbf{Mean}\\\textbf{(Type)}}} & & & \multirow{2}{*}{\parbox{1.2cm}{\centering Pair.\\Class.}} \\
Model Name & & Parameters & \textbf{Rank} &  &  & {Class.} & {Clus.} & & {Rerank.} & {Retrieval} & {STS} & {Summar.} \\
\midrule
{EmbeddingGemma (768d)} & 578 & 308M & \textcolor{blue}{\textbf{16}} & \textcolor{blue}{\textbf{69.7}} & \textcolor{blue}{\textbf{65.1}} & \textcolor{blue}{\textbf{87.6}} & \textcolor{blue}{\textbf{56.6}} & \textcolor{blue}{\textbf{87.3}} & 47.4 & 55.7 & 83.6 & \textcolor{blue}{\textbf{37.6}} \\
\hphantom{EmbeddingGemma} (512d) & & & & 69.2 & 64.6 & 87.4 & 56.5 & 87.2 & 47.2 & 54.0 & 83.6 & 36.2 \\
\hphantom{EmbeddingGemma} (256d) & & & & 68.4 & 64.0 & 86.9 & 56.2 & 86.9 & 47.0 & 51.4 & 83.8 & 35.9 \\
\hphantom{EmbeddingGemma} (128d) & & & & 66.7 & 62.7 & 85.9 & 55.9 & 86.7 & 46.9 & 46.0 & 83.3 & 34.1 \\
\midrule

GIST-large-Embedding-v0 & 1278 & 335M & 23 & 66.3 & 62.0 & 78.9 & 48.8 & 86.7 & \textbf{48.8} & 54.5 & \textbf{84.4} & 31.5\\
mxbai-embed-large-v1 & 639 & 335M & 27 & 66.3 & 62.0 & 79.1 & 47.5 & 87.2 & 48.1 & 55.4 & \textbf{84.4} & 32.6\\
UAE-Large-V1 & 1278 & 335M & 28 & 66.4 & 61.8 & 79.1 & 47.9 & 87.2 & 48.4 & \textbf{55.9} & \textbf{84.4} & 30.1\\
GIST-Embedding-v0 & 418 & 109M & 33 & 65.5 & 61.4 & 78.2 & 48.5 & 86.3 & 47.5 & 53.6 & 83.3 & 32.3\\
bge-large-en-v1.5 & 1242 & 335M & 36 & 65.9 & 61.9 & 78.3 & 48.0 & 87.1 & 48.3 & 55.4 & 82.8 & 33.1\\
GIST-small-Embedding-v0 & 127 & 33M & 41 & 64.8 & 60.7 & 78.2 & 48.0 & 84.8 & 47.3 & 52.0 & 82.8 & 31.8\\
NoInstruct-small-Embed.-v0 & 127 & 33M & 42 & - & - & - & 47.9 & 85.1 & 47.0 & 53.7 & 82.8 & 31.3\\
gte-large & 639 & 335M & 43 & 64.8 & 60.9 & 75.5 & 48.2 & 85.1 & 47.8 & 53.3 & 83.3 & 32.9\\
bge-base-en-v1.5 & 390 & 109M & 44 & 65.1 & 60.8 & 77.7 & 47.4 & 86.6 & 46.7 & 54.8 & 82.1 & 30.2\\
mini-gte & 253 & 66M & 46 & 65.1 & 60.7 & 80.0 & 47.9 & 84.8 & 46.9 & 53.2 & 81.6 & 30.3\\
SearchMap\_Preview & 1660 & 435M & 47 & 64.1 & 60.5 & 75.0 & 48.5 & 84.9 & 47.9 & 52.2 & 81.6 & 33.2\\
\bottomrule
\end{tabular}
}
\caption{Performance of top leaderboard models under 500M parameters on MTEB(Eng, v2).\\Rank is computed using Borda count against all models, regardless of parameter count.}
\label{tab:leaderboard_english}
\end{table}
\paragraph{MTEB(Eng, v2)}

Compared to the models in \Cref{tab:leaderboard_english}, EmbeddingGemma again achieves the highest overall scores and exceptional improvements in classification (+8.5), clustering (+7.8), and summarization (+4.4) compared to all other models. In fact, EmbeddingGemma also tops the sub-1B-parameter leaderboard in these task types (again excluding models with possible data leakage due to training with over 25\% of the MTEB data).

\begin{table}[!t]
\centering
\resizebox{\columnwidth}{!}{%
\begin{tabular}{lccccc|cccccccccccc}
\toprule
& \multirow{2}{*}{\parbox{1cm}{\centering Memory\\(MB)}} & & & \multirow{2}{*}{\parbox{1cm}{\centering{\textbf{Mean}\\\textbf{All}}}} & \multirow{2}{*}{\parbox{1cm}{\centering \textbf{Mean}\\\textbf{\mbox{-COIR}}}} & & & \\
Model Name & & Parameters & \textbf{Rank} &  & & {AppsR.} & {COIR} & {CESR} & {CFMT} & {CSMT}  & {CSNCCR} & {CSNR} & {CTOC} & {CTODL} & {CQA} & {SOQA} & {ST2SQL}  \\
\midrule
{EmbeddingGemma (768d)} & 578 & 308M & \textcolor{blue}{\textbf{7}} & \textcolor{blue}{\textbf{68.8}} & \textcolor{blue}{\textbf{68.1}} & \textcolor{blue}{\textbf{84.4}} & 75.5 & \textcolor{blue}{\textbf{62.1}} & 51.4 & 80.3 & 73.7 & \textcolor{blue}{\textbf{90.1}} & \textcolor{blue}{\textbf{85.5}} & 33.5 & \textcolor{blue}{\textbf{43.6}} & 86.5 & 58.4 \\
\hphantom{EmbeddingGemma} (512d) & & & & 68.5 & 67.9 & 84.3 & 74.6 & 60.9 & 50.9 & 79.8 & 73.1 & 89.7 & 85.2 & 35.0 & 44.1 & 86.0 & 58.0\\
\hphantom{EmbeddingGemma} (256d) & & & & 66.7 & 66.3 & 82.4 & 71.8 & 56.8 & 49.1 & 78.4 & 71.0 & 88.5 & 81.9 & 36.5 & 42.8 & 84.0 & 57.7\\
\hphantom{EmbeddingGemma} (128d) & & & & 63.0 & 62.7 & 79.5 & 65.9 & 50.2 & 46.1 & 76.5 & 65.7 & 86.0 & 76.0 & 36.4 & 38.5 & 79.7 & 55.0\\
\midrule
KaLM mini-v1 & 1885 & 494M & 13 & - & 63.2 & 46.8 & - & 60.0 & 52.5 & 84.0 & 59.5 & 88.0 & 79.9 & 34.0 & 33.6 & 92.4 & 64.6\\
KaLM mini-instruct-v1 & 1885 & 494M & 15 & - & 63.4 & 46.9 & - & 54.6 & 62.6 & 81.2 & 59.6 & 87.0 & 83.3 & \textbf{36.4} & 29.6 & \textbf{92.8} & 63.4\\
gte-modernbert-base & 284 & 149M & 16 & - & - & 56.4 & 83.1 & - & \textbf{85.8} & \textbf{85.5} & \textbf{93.4} & - & 73.0 & 36.1 & 42.2 & 90.9 & \textbf{64.7}\\
granite-embedding-125m-eng. & 238 & 125M & 30 & 53.3 & 53.1 & 11.8 & 55.1 & 58.8 & 42.1 & 75.3 & 47.6 & 76.9 & 66.7 & 29.6 & 36.6 & 89.8 & 48.7\\
gte-multilingual-base & 582 & 305M & 32 & - & 55.3 & 12.0 & - & 51.4 & 52.2 & 77.3 & 53.8 & 89.4 & 67.7 & 35.6 & 34.0 & 87.1 & 48.0\\
b1ade-embed & 1278 & 335M & 36 & 53.6 & 51.2 & 6.7 & \textbf{79.8} & 55.8 & 37.8 & 72.5 & 51.8 & 86.7 & 57.0 & 22.3 & 35.3 & 83.5 & 54.0\\
snowflake-arctic-embed-m-v2.0 & 1165 & 305M & 37 & - & 55.4 & 10.8 & - & 59.3 & 45.3 & 77.6 & 52.8 & 80.3 & 75.0 & 28.2 & 36.8 & 90.7 & 52.9\\
e5-large-v2 & 1278 & 335M & 38 & - & 55.3 & 14.2 & - & 57.3 & 47.8 & 76.2 & 60.0 & 83.2 & 65.1 & 32.4 & 32.1 & 89.9 & 49.7\\
granite-embedding-eng.-r2 & 284 & 149M & 40 & - & - & 14.0 & 64.7 & - & 52.5 & 77.2 & 47.7 & - & 77.1 & 35.0 & 37.0 & 91.8 & 49.6\\
granite-embedding-small-eng.-r2 & 91 & 47M & 45 & - & - & 13.5 & 60.5 & - & 52.2 & 76.9 & 48.4 & - & 77.6 & 33.6 & 35.6 & 90.0 & 46.3\\
snowflake-arctic-embed-m-long & 522 & 137M & 46 & - & 53.1 & 7.4 & - & 56.4 & 51.0 & 76.7 & 47.1 & 73.3 & 67.2 & 30.6 & 35.2 & 89.8 & 49.5\\
\bottomrule
\end{tabular}
}
\caption{Performance of top leaderboard models under 500M parameters on MTEB(Code, v1).\\Rank is computed using Borda count against all models, regardless of parameter count.}
\label{tab:leaderboard_code}
\end{table}
\paragraph{MTEB(Code)}
In \Cref{tab:leaderboard_code}, we compare results for the individual tasks present in the MTEB(Code) benchmark. Note that only a few models (including EmbeddingGemma) have been submitted to the MTEB(Code) leaderboard with evaluation metrics for all tasks. Specifically, most top models are missing COIRCodeSearchNetRetrieval (COIR); for this reason, we additionally report the mean score over the eleven remaining tasks, \textbf{Mean -COIR}. EmbeddingGemma achieves the best performance in this metric in addition to all other aggregate metrics. Observe that, compared to the second-best model, EmbeddingGemma provides dramatic performance increases in AppsRetrieval (+37.6) and CosQA (+10.0), both of which require retrieving relevant code given natural language queries—--code contest problems and web queries, respectively. This highlights EmbeddingGemma's ability to create representations which not only work across languages, but also across domains.

\begin{table}[!t]
\centering
\resizebox{\columnwidth}{!}{%
\begin{tabular}{lll|c*{20}{c}}
\toprule
& Parameters & Average & as & bho & brx & gbm & gom & gu & hi & hne & kn & mai & ml & mni & mr & mwr & or & pa & ps & sa & ta & ur \\
\midrule
{{EmbeddingGemma}} & 308M & \textcolor{blue}{\textbf{47.7}} & \textcolor{blue}{\textbf{48.4}} & \textcolor{blue}{\textbf{56.2}} & \textcolor{blue}{\textbf{12.1}} & \textcolor{blue}{\textbf{55.7}} & \textcolor{blue}{\textbf{44.7}} & \textcolor{blue}{\textbf{53.0}} & \textcolor{blue}{\textbf{63.2}} & \textcolor{blue}{\textbf{59.0}} & \textcolor{blue}{\textbf{54.0}} & \textcolor{blue}{\textbf{60.7}} & \textcolor{blue}{\textbf{55.5}} & \textcolor{blue}{\textbf{22.2}} & \textcolor{blue}{\textbf{58.3}} & \textcolor{blue}{\textbf{54.8}} & 21.2 & 45.5 & \textcolor{blue}{\textbf{35.2}} & \textcolor{blue}{\textbf{55.8}} & \textcolor{blue}{\textbf{50.3}} & \textcolor{blue}{\textbf{48.6}} \\
{{Gecko Embedding$^\dagger$}} & 278M & 7.6 & 3.2 & 10.7 & 1.0 & 12.9 & 3.7 & 2.9 & 20.4 & 11.8 & 6.9 & 13.5 & 10.4 & 1.5 & 10.3 & 12.9 & 1.5 & 1.1 & 2.5 & 8.1 & 8.5 & 8.5 \\
\midrule
gte-multilingual-base & 305M & 19.0 & 17.6 & 22.4 & 2.7 & 22.7 & 14.0 & 24.0 & 30.2 & 23.6 & 18.7 & 26.7 & 22.5 & 5.5 & 25.1 & 21.3 & 13.0 & 18.7 & 13.2 & 17.2 & 22.1 & 19.7 \\
multiling.-e5-large-instr. & 560M & 18.7 & 21.2 & 21.9 & 1.5 & 19.3 & 8.7 & 13.9 & 30.6 & 22.6 & 24.2 & 24.0 & 8.6 & 6.3 & 23.0 & 19.8 & 17.3 & 24.5 & 15.9 & 19.1 & 22.9 & 28.2 \\
bge-m3 & 568M & 26.9 & 26.5 & 31.7 & 1.8 & 29.1 & 18.7 & 31.9 & 35.1 & 31.2 & 30.3 & 32.6 & 35.2 & 5.4 & 31.8 & 29.5 & 23.1 & 29.2 & 23.4 & 30.4 & 31.0 & 30.4 \\
jina-embeddings-v3 & 572M & 8.5 & 9.8 & 7.9 & 0.1 & 7.0 & 2.8 & 12.9 & 13.8 & 7.9 & 10.1 & 8.7 & 11.5 & 0.9 & 12.2 & 8.6 & 6.7 & 11.2 & 7.0 & 8.8 & 10.7 & 12.0 \\
Qwen3-Embedding-0.6B & 595M & 6.6 & 4.6 & 9.6 & 0.3 & 8.8 & 2.1 & 6.7 & 18.1 & 9.6 & 8.0 & 10.9 & 7.0 & 0.1 & 10.2 & 8.1 & 2.0 & 8.0 & 1.0 & 7.0 & 3.7 & 6.8 \\
Linq-Embed-Mistral & 7.11B & 24.6 & 23.8 & 38.1 & 8.6 & 37.0 & 21.7 & 11.6 & 44.2 & 39.7 & 21.7 & 38.5 & 10.2 & 14.7 & 31.4 & 36.2 & 10.7 & 8.3 & 13.8 & 37.7 & 14.3 & 29.3 \\
gte-Qwen2-7B-instruct & 7.61B & 17.4 & 14.7 & 22.7 & 5.4 & 23.0 & 7.0 & 19.1 & 30.4 & 19.1 & 16.2 & 25.9 & 21.7 & 7.2 & 23.8 & 24.0 & 11.3 & 19.2 & 11.0 & 21.1 & 9.7 & 15.5 \\
voyage-3-large & -- & 39.2 & 34.3 & 44.8 & 7.9 & 46.6 & 27.1 & 46.7 & 54.3 & 45.3 & 41.5 & 48.3 & 45.3 & 19.2 & 45.5 & 47.9 & \textbf{32.3} & \textbf{48.4} & 26.8 & 40.0 & 36.0 & 45.6 \\
text-embedding-3-large & -- & 18.8 & 18.2 & 28.8 & 3.3 & 28.4 & 11.1 & 14.6 & 40.4 & 29.3 & 17.1 & 31.1 & 15.6 & 2.9 & 25.5 & 28.7 & 8.3 & 11.3 & 6.8 & 26.6 & 6.0 & 22.0 \\
\bottomrule
\end{tabular}
}
\caption{Performance of top multilingual models on XTREME-UP (MRR@10). $^\dagger$: For Gecko Embedding~\citep{lee2024gecko}, we evaluate a smaller version of the model discussed in the paper.}
\label{tab:leaderboard_xtreme}
\end{table}
\paragraph{XTREME-UP}
\Cref{tab:leaderboard_xtreme} shows results on XTREME-UP for EmbeddingGemma and other top multilingual models. The XTREME-UP cross-lingual retrieval task involves mapping queries in 20 underrepresented languages to English passages. EmbeddingGemma's performance is remarkable, vastly outperforming top models with billions of parameters as well as commercial API models. Moreover, against the selected open models, EmbeddingGemma achieves the strongest performance for each individual language evaluated, being outperformed only by a commercial model in two languages. Note that even models with strong performance in the MTEB multilingual benchmark may struggle in XTREME-UP, such as Qwen3 Embedding 0.6B \citep{zhang2025qwen3embeddingadvancingtext} and Jina Embeddings v3 \citep{sturua2024jinaembeddingsv3multilingualembeddingstask}. This highlights EmbeddingGemma's exceptional capability in low-resource languages.

\section{Future Work}
We intend to extend EmbeddingGemma's capabilities beyond text, into modalities such as image, audio, and video. This includes exploring new recipes to excel simultaneously at various unimodal (e.g. text $\Leftrightarrow$ text), cross-modal (e.g. text $\Leftrightarrow$ image), and multimodal (e.g. text+image $\Leftrightarrow$ image) use cases. Gemma 3 has demonstrated strong multimodal understanding, and previous work has shown promise in adapting multimodal large language models into multimodal embedding models, (\citeauthor{e5v}, \citeyear{e5v}; \citeauthor{vlm2vec}, \citeyear{vlm2vec}). However, these models use several billion parameters; we aim to fulfill the need for lightweight, natively multimodal embedding models able to be run on-device.

\section{Conclusion}
EmbeddingGemma is a lightweight yet powerful general-purpose embedding model built upon the Gemma 3 language model family. Our novel training recipe leverages knowledge from larger models through initialization from a rich T5Gemma encoder and distillation from the highly-capable Gemini Embedding model. It also includes spread-out regularization and model merging with multiple optimized mixtures, to ensure that the resulting representations are expressive and generalizable. As a result, EmbeddingGemma sets a new standard for resource-efficient text embeddings.

Our extensive evaluations demonstrate that EmbeddingGemma pushes the state-of-the-art across multilingual, English, and code benchmarks on the Massive Text Embedding Benchmark (MTEB) leaderboard for models under 500M parameters, and its performance is even comparable to models twice its size. This lead persists when using quantization for low-cost inference or truncating embeddings to reduce storage costs.

By delivering state-of-the-art performance in a compact size, EmbeddingGemma addresses the growing demand for models that enable faster, private, and offline-capable applications directly on user devices. We release our checkpoints to the community to accelerate research and development in efficient models.

\bibliography{main}
\appendix
\newpage
\section{Full Results}

\begin{table}[ht]
\centering
\resizebox{0.3435\columnwidth}{!}{%
\begin{tabular}{lc}
\toprule
Task Name & Performance \\
\midrule
AILAStatutes & 37.37 \\
AfriSentiClassification & 44.47 \\
AlloProfClusteringS2S.v2 & 52.82 \\
AlloprofReranking & 79.69 \\
AmazonCounterfactualClassification & 84.23 \\
ArXivHierarchicalClusteringP2P & 63.59 \\
ArXivHierarchicalClusteringS2S & 59.59 \\
ArguAna & 71.54 \\
ArmenianParaphrasePC & 92.68 \\
BUCC.v2 & 98.75 \\
BelebeleRetrieval & 72.38 \\
BibleNLPBitextMining & 12.68 \\
BigPatentClustering.v2 & 41.64 \\
BiorxivClusteringP2P.v2 & 52.10 \\
BornholmBitextMining & 34.63 \\
BrazilianToxicTweetsClassification & 22.34 \\
BulgarianStoreReviewSentimentClassfication & 71.26 \\
CEDRClassification & 52.76 \\
CLSClusteringP2P.v2 & 41.45 \\
CSFDSKMovieReviewSentimentClassification & 34.49 \\
CTKFactsNLI & 79.34 \\
CataloniaTweetClassification & 51.23 \\
Core17InstructionRetrieval & 6.31 \\
CovidRetrieval & 78.93 \\
CyrillicTurkicLangClassification & 58.63 \\
CzechProductReviewSentimentClassification & 58.63 \\
DBpediaClassification & 94.27 \\
DalajClassification & 50.28 \\
DiaBlaBitextMining & 83.93 \\
EstonianValenceClassification & 38.29 \\
FaroeseSTS & 65.30 \\
FilipinoShopeeReviewsClassification & 40.52 \\
FinParaSTS & 25.22 \\
FinancialPhrasebankClassification & 86.45 \\
FloresBitextMining & 55.35 \\
GermanSTSBenchmark & 84.67 \\
GreekLegalCodeClassification & 29.03 \\
GujaratiNewsClassification & 82.78 \\
HALClusteringS2S.v2 & 29.34 \\
HagridRetrieval & 98.92 \\
IN22GenBitextMining & 74.40 \\
IndicCrosslingualSTS & 43.07 \\
IndicGenBenchFloresBitextMining & 87.09 \\
IndicLangClassification & 46.62 \\
IndonesianIdClickbaitClassification & 60.87 \\
IsiZuluNewsClassification & 26.42 \\
ItaCaseholdClassification & 70.36 \\
JSICK & 84.40 \\
KorHateSpeechMLClassification & 11.58 \\
KorSarcasmClassification & 58.10 \\
KurdishSentimentClassification & 59.97 \\
LEMBPasskeyRetrieval & 60.75 \\
LegalBenchCorporateLobbying & 95.08 \\
MIRACLRetrievalHardNegatives & 66.20 \\
MLQARetrieval & 78.95 \\
MacedonianTweetSentimentClassification & 45.31 \\
MalteseNewsClassification & 33.08 \\
MasakhaNEWSClassification & 74.93 \\
MasakhaNEWSClusteringS2S & 43.46 \\
MassiveIntentClassification & 62.70 \\
MedrxivClusteringP2P.v2 & 44.11 \\
MultiEURLEXMultilabelClassification & 4.34 \\
MultiHateClassification & 61.00 \\
NTREXBitextMining & 73.87 \\
NepaliNewsClassification & 95.48 \\
News21InstructionRetrieval & 11.45 \\
\bottomrule
\end{tabular}}
\hspace{1.5em}
\centering
\resizebox{0.342\columnwidth}{!}{%
\begin{tabular}{lc}
\toprule
Task Name & Performance \\
\midrule
NollySentiBitextMining & 41.26 \\
NordicLangClassification & 65.56 \\
NorwegianCourtsBitextMining & 90.79 \\
NusaParagraphEmotionClassification & 44.16 \\
NusaTranslationBitextMining & 66.05 \\
NusaX-senti & 69.80 \\
NusaXBitextMining & 67.11 \\
OdiaNewsClassification & 57.95 \\
OpusparcusPC & 93.35 \\
PAC & 67.87 \\
PawsXPairClassification & 57.73 \\
PlscClusteringP2P.v2 & 72.14 \\
PoemSentimentClassification & 58.86 \\
PolEmo2.0-OUT & 62.77 \\
PpcPC & 90.86 \\
PunjabiNewsClassification & 82.36 \\
RTE3 & 89.67 \\
Robust04InstructionRetrieval & -0.94 \\
RomaniBibleClustering & 41.88 \\
RuBQReranking & 71.26 \\
SCIDOCS & 18.43 \\
SIB200ClusteringS2S & 26.53 \\
SICK-R & 81.37 \\
STS12 & 79.32 \\
STS13 & 86.42 \\
STS14 & 83.67 \\
STS15 & 89.35 \\
STS17 & 84.42 \\
STS22.v2 & 71.20 \\
STSB & 81.64 \\
STSBenchmark & 88.16 \\
STSES & 82.31 \\
ScalaClassification & 50.77 \\
SemRel24STS & 65.22 \\
SentimentAnalysisHindi & 65.49 \\
SinhalaNewsClassification & 65.67 \\
SiswatiNewsClassification & 57.00 \\
SlovakMovieReviewSentimentClassification & 73.27 \\
SpartQA & 10.68 \\
SprintDuplicateQuestions & 97.03 \\
StackExchangeClustering.v2 & 90.94 \\
StackOverflowQA & 86.47 \\
StatcanDialogueDatasetRetrieval & 46.27 \\
SwahiliNewsClassification & 65.95 \\
SwednClusteringP2P & 40.04 \\
SwissJudgementClassification & 57.74 \\
T2Reranking & 67.54 \\
TERRa & 65.15 \\
TRECCOVID & 80.35 \\
Tatoeba & 51.35 \\
TempReasonL1 & 1.00 \\
ToxicConversationsClassification & 82.93 \\
TswanaNewsClassification & 31.15 \\
TweetTopicSingleClassification & 73.02 \\
TwitterHjerneRetrieval & 72.04 \\
TwitterURLCorpus & 86.90 \\
VoyageMMarcoReranking & 60.99 \\
WebLINXCandidatesReranking & 10.16 \\
WikiCitiesClustering & 92.02 \\
WikiClusteringP2P.v2 & 27.03 \\
WikipediaRerankingMultilingual & 89.88 \\
WikipediaRetrievalMultilingual & 90.00 \\
WinoGrande & 59.40 \\
XNLI & 81.73 \\
indonli & 60.95 \\
\bottomrule
\end{tabular}
}
\caption{Full results of EmbeddingGemma on MTEB(Multilingual, v2).}
\end{table}
\begin{table}[t!]
\label{sec:full_results_scatter}
\centering
\begin{minipage}[t]{0.3435\columnwidth}
\centering
\resizebox{\columnwidth}{!}{%
\begin{tabular}{lc}
\toprule
Task Name & Performance \\
\midrule
AmazonCounterfactualClassification & 90.07 \\
ArXivHierarchicalClusteringP2P & 63.59 \\
ArXivHierarchicalClusteringS2S & 59.59 \\
ArguAna & 71.54 \\
AskUbuntuDupQuestions & 62.95 \\
BIOSSES & 86.38 \\
Banking77Classification & 91.45 \\
BiorxivClusteringP2P.v2 & 52.10 \\
CQADupstackGamingRetrieval & 59.52 \\
CQADupstackUnixRetrieval & 41.52 \\
ClimateFEVERHardNegatives & 26.71 \\
FEVERHardNegatives & 80.75 \\
FiQA2018 & 47.74 \\
HotpotQAHardNegatives & 71.48 \\
ImdbClassification & 92.92 \\
MTOPDomainClassification & 99.11 \\
MassiveIntentClassification & 85.79 \\
MassiveScenarioClassification & 91.54 \\
MedrxivClusteringP2P.v2 & 44.11 \\
MedrxivClusteringS2S.v2 & 41.93 \\
MindSmallReranking & 31.90 \\
SCIDOCS & 18.43 \\
SICK-R & 81.37 \\
STS12 & 79.32 \\
STS13 & 86.42 \\
STS14 & 83.67 \\
STS15 & 89.35 \\
STS17 & 90.31 \\
STS22.v2 & 67.52 \\
STSBenchmark & 88.16 \\
SprintDuplicateQuestions & 97.03 \\
StackExchangeClustering.v2 & 90.94 \\
StackExchangeClusteringP2P.v2 & 48.90 \\
SummEvalSummarization.v2 & 37.64 \\
TRECCOVID & 80.35 \\
Touche2020Retrieval.v3 & 58.90 \\
ToxicConversationsClassification & 82.93 \\
TweetSentimentExtractionClassification & 66.59 \\
TwentyNewsgroupsClustering.v2 & 51.29 \\
TwitterSemEval2015 & 77.94 \\
TwitterURLCorpus & 86.90 \\
\bottomrule
\end{tabular}}
\end{minipage}
\quad
\begin{minipage}{0.495\textwidth}
\centering
\resizebox{0.65\columnwidth}{!}{%
\begin{tabular}{lc}
\toprule
Task Name & Performance \\
\midrule
AppsRetrieval & 84.39 \\
COIRCodeSearchNetRetrieval & 75.54 \\
CodeEditSearchRetrieval & 62.10 \\
CodeFeedbackMT & 51.42 \\
CodeFeedbackST & 80.26 \\
CodeSearchNetCCRetrieval & 73.71 \\
CodeSearchNetRetrieval & 90.15 \\
CodeTransOceanContest & 85.51 \\
CodeTransOceanDL & 33.52 \\
CosQA & 43.60 \\
StackOverflowQA & 86.46 \\
SyntheticText2SQL & 58.42 \\
\bottomrule
\end{tabular}
}
\par\vspace{1em}
\includegraphics[width=0.95\columnwidth]{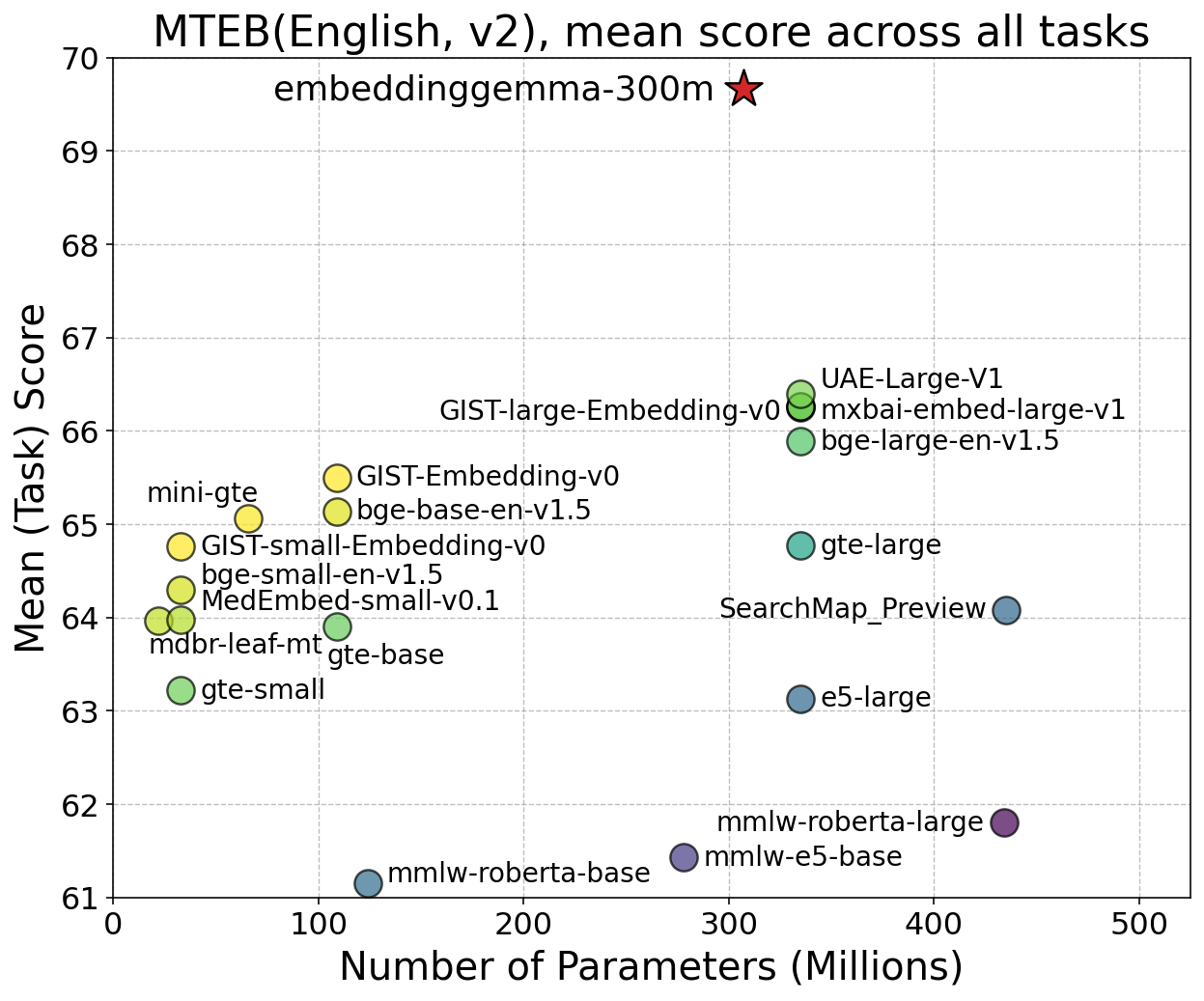}
\end{minipage}
\caption{Full results of EmbeddingGemma on MTEB(English, v2) (left) and MTEB(Code) (right). We include the comparison of top 20 embedding models under 500M parameters in MTEB(English, v2), excluding models trained on more than 25\% of the MTEB data to mitigate potential over‑fitting.}
\end{table}
\begin{table}[ht]
\centering
\resizebox{0.35\columnwidth}{!}{%
\begin{tabular}{l r c c}
\toprule
& & \multicolumn{2}{c}{Performance} \\
\cmidrule(lr){3-4}
\multicolumn{2}{c}{Language} & R@5000kt & R@2000kt \\
\midrule
Arabic & (ar) & 83.50 & 75.08 \\
Bengali & (bn) & 91.45 & 85.86 \\
Finnish & (fi) & 78.34 & 71.97 \\
Japanese & (ja) & 81.33 & 72.61 \\
Korean & (ko) & 83.86 & 78.25 \\
Russian & (ru) & 82.70 & 78.90 \\
Telugu & (te) & 87.82 & 83.19 \\
\bottomrule
\end{tabular}
}
\quad\quad\quad
\centering
\resizebox{0.3\columnwidth}{!}{%
\begin{tabular}{lrc}
\toprule
& & Performance \\
\multicolumn{2}{c}{Language} & (MRR@10) \\
\midrule
Assamese & (as) & 48.44 \\
Bhojpuri & (bho) & 56.21 \\
Boro & (brx) & 12.14 \\
Garhwali & (gbm) & 55.71 \\
Konkani & (gom) & 44.65 \\
Gujrati & (gu) & 52.95 \\
Hindi & (hi) & 63.19 \\
Chhattisgarhi & (hne) & 58.96 \\
Kannada & (kn) & 54.04 \\
Maithili & (mai) & 60.74 \\
Malayalam & (ml) & 55.46 \\
Manipuri & (mni) & 22.16 \\
Marathi & (mr) & 58.27 \\
Marwari & (mwr) & 54.81 \\
Odia & (or) & 21.21 \\
Punjabi & (pa) & 45.50 \\
Pashto & (ps) & 35.21 \\
Sanskrit & (sa) & 55.81 \\
Tamil & (ta) & 50.34 \\
Urdu & (ur) & 48.61 \\
\bottomrule
\end{tabular}
}
\caption{Full results of EmbeddingGemma on XOR-Retrieve (left) and XTREME-UP (right).}
\end{table}

\twocolumn[  
    \begin{@twocolumnfalse}
        \vspace{-1.24em}
        \section{Contributions and Acknowledgments}
     \end{@twocolumnfalse}
]

\enlargethispage{\baselineskip}

\textbf{Core Contributors} ($^*$: equal contributions)\\
Henrique Schechter Vera$^*$\\
Sahil Dua$^*$\\
Daniel Salz \\
Ryan Mullins \\
Sindhu Raghuram Panyam \\
Sara Smoot \\
Iftekhar Naim \\
Joe Zou \\
Feiyang Chen \\
Daniel Cer\vspace{1em}\\
\textbf{Contributors}\\
Alice Lisak \\
Min Choi \\
Lucas Gonzalez \\
Omar Sanseviero \\
Glenn Cameron \\
Ian Ballantyne \\
Kat Black \\
Kaifeng Chen \\
Weiyi Wang \\
Zhe Li \\
Gus Martins \\
Jinhyuk Lee \\
Mark Sherwood \\
Juyeong Ji \\
Renjie Wu \\
Jingxiao Zheng \\
Jyotinder Singh \\
Abheesht Sharma \\
Divyashree Sreepathihalli \vspace{0.5em}\\
\noindent\textbf{Gemini Embedding} \textit{(alphabetical order)}\\
Aashi Jain \\
Adham Elarabawy \\
AJ Co \\
Andreas Doumanoglou \\
Babak Samari \\
Ben Hora \\
Brian Potetz \\
Dahun Kim \\
Enrique Alfonseca \\
Fedor Moiseev \\
Feng Han \\
Frank Palma Gomez \\
Gustavo Hern{\'{a}}ndez {\'{A}}brego \\
Hesen Zhang \\
Hui Hui \\
Jay Han \\
Karan Gill \\
Ke Chen \\
Koert Chen \\
Madhuri Shanbhogue \\
Michael Boratko \\
Sai Meher Karthik Duddu \\
Sandeep Mariserla \\
Setareh Ariafar \\
Shanfeng Zhang \\
Shijie Zhang \\
Simon Baumgartner \\
Sonam Goenka \\
Steve Qiu \\
Tanmaya Dabral \\
Trevor Walker \\
Vikram Rao \\
Waleed Khawaja \\
Wenlei Zhou \\
Xiaoqi Ren \\
Ye Xia \\
Yichang Chen \\
Yi-Ting Chen \\
Zhe Dong \\
Zhongli Ding \vspace{0.5em} \\
\noindent\textbf{Gemma} \textit{(alphabetical order)}\\
Francesco Visin \\
Jiageng Zhang \\
Kathleen Kenealy \\
Michelle Casbon \\
Ravin Kumar \\
Thomas Mesnard \vspace{0.5em}\\
\noindent\textbf{T5Gemma} \textit{(alphabetical order)}\\
Biao Zhang \\
Ga\"{e}l Liu \\
Paul Suganthan \vspace{0.5em}\\
\noindent\textbf{Leadership}\\
Zach Gleicher \\
Cormac Brick \\
Olivier Lacombe \\
Adam Roberts \\
Qin Yin \\
Yunhsuan Sung \\
Raphael Hoffmann \\
Tris Warkentin \\
Armand Joulin \\
Tom Duerig \\
Mojtaba Seyedhosseini

\onecolumn
\noindent\textbf{Acknowledgement}\\
Daniel Estrada Alva, Krzysztof Czuba, Howard Zhou, Tom\'{a}\v{s} I\v{z}o, Tom Aarsen, Aditya Kusupati, Samer~Hassan, Aishwarya Nagarajan, Imed Zitouni

\end{document}